\documentclass[10pt,journal,compsoc]{IEEEtran}

\usepackage[noadjust]{cite}
\usepackage{filecontents}

\usepackage{graphics}
\usepackage{lipsum}
\usepackage{graphicx,amsmath,alltt}
\usepackage[figure,linesnumbered]{algorithm2e}
\usepackage{textcomp}
\usepackage{booktabs}
\usepackage{multirow}
\usepackage{colortbl}
\usepackage{ragged2e}
\usepackage{tabularx,booktabs}
\usepackage[dvipsnames,table,xcdraw]{xcolor}
\renewcommand{\citedash}{--}

\begin{document}


\title{
\normalfont{Technical Report} \\
\normalfont\Large\bfseries{Double Q-Learning for Citizen Relocation During Natural Hazards}
}

\author{
    \IEEEauthorblockN {
       Alysson~Ribeiro~da~Silva
    }
   
    \IEEEauthorblockA {
        Computer Science Graduate Program \\
        Federal University of Minas Gerais \\
        Belo Horizonte, Brazil
    }
}

\IEEEtitleabstractindextext{%


\justify
\begin{abstract}
Natural disasters can cause substantial negative socio-economic impacts around
the world, due to mortality, relocation, rates, and reconstruction decisions.
Robotics has been successfully applied to identify and rescue victims
during the occurrence of a natural hazard. However, little effort has been
taken to deploy solutions where an autonomous robot can save the life of
a citizen by itself relocating it, without the need to wait for a rescue team
composed of humans. Reinforcement learning approaches can be used to
deploy such a solution, however, one of the most famous algorithms to deploy
it, the Q-learning, suffers from biased results generated when performing its
learning routines. In this research a solution for citizen relocation based on
Partially Observable Markov Decision Processes is adopted, where the capability
of the Double Q-learning in relocating citizens during a natural hazard is
evaluated under a proposed hazard simulation engine based on a grid world.
The performance of the solution was measured as a success rate of a citizen
relocation procedure, where the results show that the technique portrays a
performance above 100\% for easy scenarios and near 50\% for hard ones.
\end{abstract}


}

\maketitle
\IEEEdisplaynontitleabstractindextext
\IEEEpeerreviewmaketitle
\ifCLASSOPTIONcompsoc


\section{Introduction}

Natural disasters can cause substantial negative socio-economic impacts around the world, due to mortality, relocation, rates, and reconstruction decisions. In Japan, the earthquake that subsequently provoked a tsunami and nuclear disaster in 2011 and the Nepal Gorkha earthquake in 2015 caused human life losses, recovery, and post-hazard reconstruction expenses that those countries are yet facing due to the affected area size and population \cite{2017_Editorial_Nature}. River floods are also sources of natural hazards, since vulnerabilities in developed and underdeveloped countries around the world increase and decrease according to the region and its economy \cite{2016_Tanoue_Nature}. Relocation and reconstruction after a natural hazard are crucial to resettle socio-economical growth. However, studies show that the reconstructed area is often not adapted to reduce the risks from a natural hazard. Besides that, poor families tend to relocate directly to the area where a hazard occurred, due to its lower price if compared to areas not affected by it \cite{2018_McCaughey_Nature}. Despite developed and other counties being able to pay for vulnerability fixes, they do not tend to explore all available technology to help spare human lives and avoid social damage, which can potentially reduce the overall impact caused by a natural hazard.

Robotics has been successfully applied to identify and rescue victims during the occurrence of a natural hazard. Post-hazard solutions such as Aerial drones can be used to help rescue victims during tsunamis, floods, earthquakes, and volcano eruptions \cite{2016_Lee_IEEEConference}. Moreover, communication with a human-based rescue team, during a hazard, is essential to provide an environment to facilitate the rescue operation \cite{2017_Baumgartner_IEEEConference}. Efforts of creating a communication network that can previously turn the citizens aware of the coming hazard are also taken in some countries, but they do not help citizens to relocate in time before the hazard reaches them or even during its occurrence. Autonomous robots can potentially save lives, by carrying citizens to safe locations with their path planning algorithms, consequently reducing the damage costs to a country’s infrastructure and economy. However, little effort has been taken to deploy solutions where an autonomous robot can save the life of a citizen by itself, without the need to wait for a rescue team composed of humans, even in countries that are able to pay for it. Due to that fact, this research aims at proposing path planning algorithms to be deployed for citizen rescue during or before the occurrence of natural hazards.

An autonomous robot, that can relocate citizens to a safe place, needs to navigate through a changing environment, where debris and other objects can influence its trajectory and the safety of the person being rescued. This situation describes the path planning under motion uncertainty problem, where states are imperfect and lack the necessary information to perform conventional algorithms, such as the uniform-cost search and its non-stochastic variants. 

In order to tackle path planning under uncertainty problems, linear-quadratic Gaussian motion planning can be used \cite{2011_Berg}. It is based on a linear-quadratic controller with a Gaussian model of uncertainty that helps to evaluate the quality of the generated path. Similar models were also applied for motion planning in reconfigurable robots to ensure more conservative poses under uncertainties \cite{2016_Norouzi}. Moreover, this approach can also be adapted to build paths while a robot is doing its sensing in environments that present uncertainties \cite{2017_Pilania}. Since generated paths need to be safe, to transport citizens, it is also important to provide safer navigation in three-dimensional environments. This task is accomplished by \cite{2018_Lunenburg} by checking the collision probability in a volumetric space. Many of the solutions tend to use a stochastic approach to deal with the uncertainty of the environment. None of the cited works deals with the uncertainties with a learning technique such as Reinforcement learning.

Reinforcement learning approaches are also well present in the literature, however, one of the most famous algorithms to deploy it, the Q-learning, suffers from biased results when performing its learning routines. In this research, a stochastic motion planner is also explored. However, different from the aforementioned researches and considerations, it is based on Partially Observable Markov Decision Processes, where the capability of the Double Q-learning in relocating citizens during a natural hazard is explored. This technique is used in this research to conduct the behavior of the agent and also to avoid biased decisions.
\section{Temporal Learning Through Double Q-Learning}

A temporal learning method that have being used to help a machine learn, in reinforcement learning, is called Double Q-Learning. It is used in this research as insurance, in case of critical failures, to help the autonomous robot to relocate citizens under uncertainties since the common Q-Learning method can produce biased results in terms of action selection through the $\arg\max$ policy.

The dynamics from the Double Q-Learning are governed by Equations \ref{eq:double_q_learning} and \ref{eq:double_q_learning_U}, where $s$ is the current state, $a$ an action that will be executed upon the presentation of $s$, $Q(a,b)$ is a virtual value, stored inside a structure called Q-table $Q$, that represents how much reinforcement is being registered to the state $a$ after performing the action $b$, $r$ is the immediate reward received after performing $a$, and $\arg\max_{a'}Q(s',a')$ being the best virtual reinforcement value from the next state, $s'$ after performing the action $a'$. The variables, $\alpha$, and $\gamma$ are used to determine the learning rate and a discount parameter, respectively. To reduce biased Q-values, a second virtual Q-table $Q^u$, storing all $Q^u(s,a)$, is used. The two Q-tables, $Q$, and $Q^u$ are interleaved at each iteration to learn,

\begin{equation}\label{eq:double_q_learning}
\begin{split}
a* & = \arg \max_{a'}Q(s',a') \\
Q(s,a) & = Q(s,a) + \alpha [r + \arg \max_{a*}Q^u(s',a*) - Q(s,a)]
\end{split}
\end{equation}

\begin{equation}\label{eq:double_q_learning_U}
\begin{split}
a* & = \arg \max_{a'}Q^u(s',a') \\
Q^u(s,a) & = Q^u(s,a) + \alpha [r + \arg \max_{a*}Q(s',a*) - Q^u(s,a)] 
\end{split}
\end{equation}

\noindent
where the predicted action by $\arg \max_{a*}$ is obtained by the adjacent Q-table, if updating $Q$ it is obtained from $Q^u$ and reciprocally for $Q^u$.


The double Q-Learning method is performed by an agent consecutive times, storing its virtual Q values, represented by the function $Q(s,a)$, until reaching a convergence, or until its values do not change over an observation period. To help keep track of boundaries, when performing the Q-learning, methods such as the boundary Q-Learning, as described by Equation \ref{eq:boundary}, which is using the self scale factor $1 - Q(s,a)$ to keep the $TDerr$ inside of the range $[0, 1]$, can be used since it helps normalize the Q-values inside a standardized working interval.

\begin{equation} \label{eq:boundary}
\begin{split}
TDerr & = \alpha [r + \gamma \arg\max_{a*} Q^u(s',a*) - Q(s,a)] \\ 
TDerr & = (1 - Q(s,a))TDerr \\
Q(s,a) & = Q(s,a) + TDerr
\end{split}
\end{equation}

To select the proper action when performing the Q-Learning, an action selection policy is used and it allows to tell the algorithm when to explore, select a random action to see what will happen, and exploit, and select sub-optimal formed strategies among all the stored virtual Q-values. In this research, the utilized action selection policy is the $\epsilon-Grreedy$, since it is simple and easy to implement. It is based on a threshold value $\epsilon$, where its dynamics are described by Equation \ref{eq:egreedy},

\begin{equation}\label{eq:egreedy}
a = \begin{cases}
\arg \max_{a'}Q(s,a) & \text{if $var > \epsilon$}\\
select random a & \text{otherwise}
\end{cases}  
\end{equation}

\noindent
where $var$ is a random value generated between the interval $[0,1]$ and $\epsilon$ is decremented after each iteration of the algorithm by a decay factor.
\section{Method}

The proposal consists of the deployment of autonomous segway robots, that can carry people to a chosen destination during the occurrence of a natural hazard. Figure \ref{fig:method.png} is depicted as part of Osaka city, in Japan, where the blue arrows symbolize a possible path from which tsunamis can arise, the orange circle represents an agent, and the green part of the picture a safe zone. In this scenario, the solution, implanted in a segway robot, receives a signal from an outsource, an observation station, telling from where the hazard will most likely start to affect the city. From there, assuming that there is a critical failure in its sensor, preventing it from fully observing the environment and performing deterministic approaches, it turns on the reinforcement learning switch. With the switch turned on, the robot navigates through the environment, with previously learned data from training sessions, and brings a citizen to a safe location.

The main process of the proposal is a three-phase routine described as follows.

\begin{itemize}
	\item \textbf{Perception}: This phase is responsible to tell the robot whether a natural hazard is happening or not. This activates sensors on it to call for a passenger that is in its vicinity.
	\item \textbf{Action}: During this phase, a passenger is already on top of the robot and is being relocated through the usage of deterministic approaches.
	\item \textbf{Insurance}: This is the phase, where the reinforcement learning is activated due to eventual critical system failures that may result in the robot being unable to fully sense its surroundings.
\end{itemize}

\noindent
For instance, in Figure \ref{fig:method.png}, the \textit{Perception} phase is activated right after the event and the robot is yet located in its initial position, the \textit{Action} phase, on the other hand, is happening while the robot is relocating a passenger through an optimal path found by a deterministic approach and represented by the orange line, finally, the \textit{Insurance} phase is turned on after a critical failure moment, represented by the vertical red line, and is responsible to finish conducting the passenger to the save zone on the right of the map depicted as a green area.  

\begin{figure}[ht]
    \centering
    \includegraphics[width=0.45\textwidth]{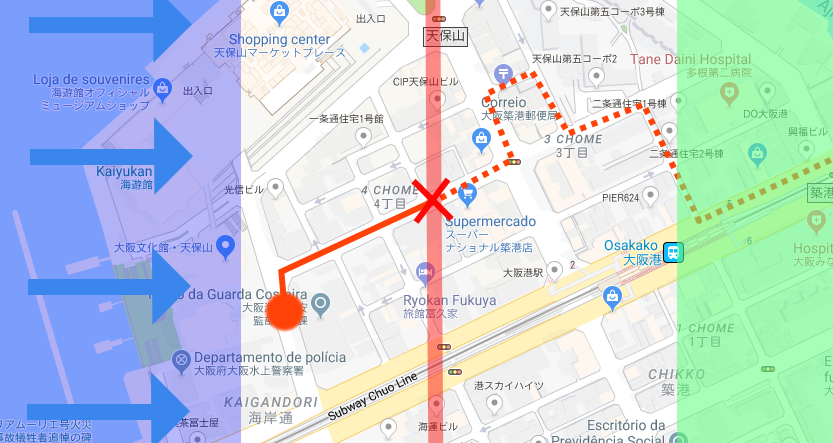}
    \caption{Citizen relocation path generated by an agent during a hypothetical natural hazard, depicted by the blue area and arrows, relocating to a safe zone, depicted as the green area on the right.}
    \label{fig:method.png}
\end{figure}

\section{Experimental Setup}

In order to test the viability of the Double Q-Learning method in relocating the citizens to a safe location during the natural hazard, it was developed a hazard simulation engine based on a grid environment. This engine simulates a $w \times h$ grid world, where $w$ is its width and $h$ its height. Each position of this grid world contains a symbol used to define a hazard area, a free space, or an obstacle. As an example of a grid world, depicted in Figure \ref{fig:gridworld}, the characters $@$, $.$, and $\#$, are representing an obstacle zone, a free path, and a hazard area, respectively.

\begin{figure}
	\label{fig:gridworld}
	\centering
	\includegraphics[width=0.45\textwidth]{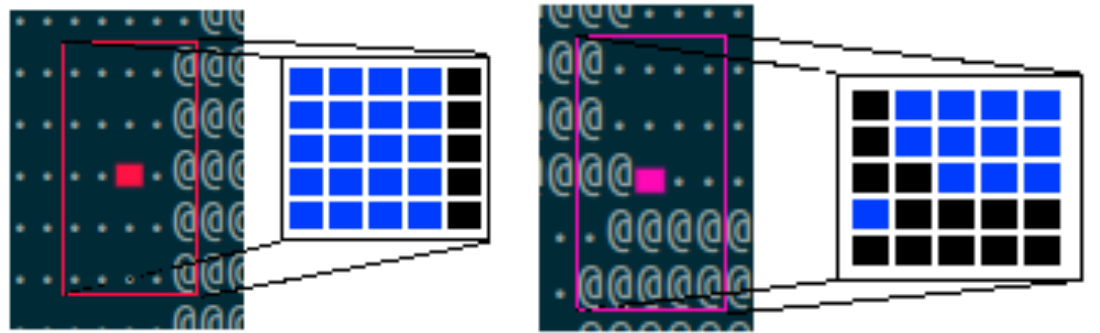}
	\caption{Grid world example, with obstacles depicted as $@$, walking zones depicted as $.$, and hazard areas, depicted as $\#$. The agent state is the region of the map depicted by the blue squares.}
\end{figure}

In order to simulate hazards, such as tsunamis, volcano eruptions, and floods, a hazard mechanism was deployed into the engine. This flooding mechanism allows for the simulation of different types of hazards in real-time, at the same time the agent is performing in the grid world, where they manipulate the location, movement, and velocity of the hazarded areas in the grid world. Five types of flooding simulations were added to the first version of the engine, depicted in Figure \ref{fig:flooding_types}. The first type, depicted by the square with the letter \textit{A}, is a ping, that grows from a starting point with a radius $r$ that is incremented by a $\delta r$ factor over time. On the other hand, the types depicted by the \textit{B} and \textit{C} squares are also pings, however, their origin is always the top right or bottom right corner of the grid world. Next, the letter \textit{D} represents vertical flooding that is increased $\delta d$ cells over time. Finally, the last flooding method portrays a catastrophic situation in which various pings are placed randomly over the grid world. This last method in special, can pop several pings at once or stay several cycles without portraying a hazard.

\begin{figure}
	\label{fig:flooding_types}
	\centering
	\includegraphics[width=0.45\textwidth]{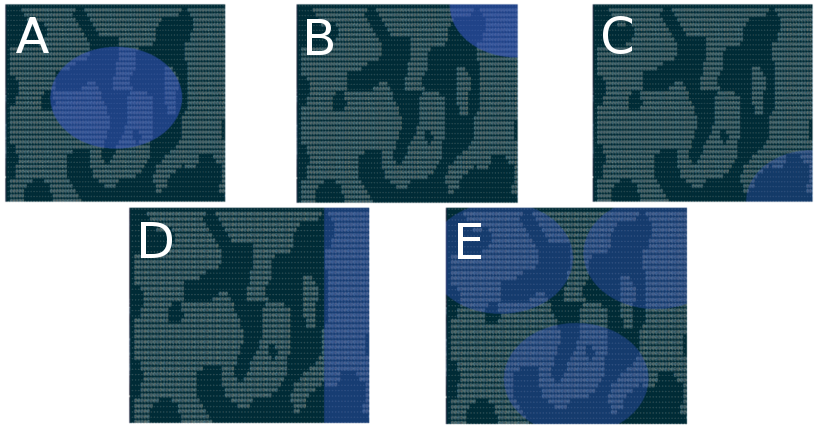}
	\caption{Flooding engine flooding types. The \textbf{A} flooding represents a ping, \textbf{B} and \textbf{C} represents an upper and bottom right pings, respectively, \textbf{D} represents a vertical flooding, and \textbf{E} represents a randomized ping flooding scenario.}
\end{figure}

The agent that acts in the proposed grid world hazard environment possesses a deliberative real-time architecture and it is equipped with two sensors, a proximity sensor, and a binary sonar. The proximity sensor is capable to sense an $8 \times 8$ feature map $F$ around itself, where at each iteration of the engine, it updates its values. Each position $f_i \in F$ stores a ternary variable that is $0$ when colliding with an empty space, $1$ for an obstacle, and $2$ for a hazarded area. Moreover, the binary sonar can detect if there is an obstacle surrounding the agent, in a cross formation, at a certain distance measured with integers that are stored in a four-position sensor array $S$. It is important to note that the agent does not possess any explicit positioning mechanism, that can inform to it where it is currently located on the map, and it also does have not a mechanism that allows it to sense exactly where a hazarded area is located. The only information that it possesses besides its feature map and binary sensors is a vector array that tells which area of the map may be affected by hazards at a time $t$ during the simulation. For the agent, a full state observed during the time $t$ is a vector $E$ composed by each $f_i$ and each $s_i \in S$.

In order to deploy the Double Q-Learning, two dynamic Q-Tables, that grow according to the necessity and due to how many different states are being observed by the agent, were implemented. The dynamic Q-Table was preferred over other structures since it reduces the computational cost to access and store Q-values. Each entry of the table is indexed by a string key and its content stores a full observed state $E$ alongside the performed action $a$ and the received reward for that action. For instance, the agent can perform $8$ actions, representing movements in its vicinity, portrayed by a $3\times3$ cell mask.

\section{Experimental Design}

The Double Q-Learning was evaluated in terms of success rate, measured as the success count over the total amount of performed simulations. In order to evaluate its success rate, $20$, $10$ with fewer obstacles and $10$ plenty with obstacles, maps were generated for the grid sizes $32\times32$, $64\times64$, and $128\times128$. Only one agent was tested in the world, where its initial position is always located on the right and the safe zone points are always located on the left of the map. To consider statistical fluctuations, it was generated $1000$ random points as starting positions and $1000$ as safe zones, where $100$ are chosen randomly, to serve as a standard benchmark. The historic success rate of the agent is computed as the average of $50$ iterations of all maps for all map sizes, called epochs, over a total of $1000$ iterations. To verify the impact of the statistical fluctuations provided by the environment, this experiment is repeated $1000$ times varying randomly the start and ending points of the agent, where more than $100$ million simulations were performed in total. 

\section{Results}

As depicted in Figure \ref{fig:32x32}, when performing in a less dense environment, the algorithm obtained the best performance when facing a central ping flooding, where it reached a $100\%$ success rate. Its worst performance occurred when facing linear flooding, near a $20\%$ success rate. The randomized ping flooding made the agent develop $60\%$ performance. In the dense environment, on the other hand, was observed a drop, near $10\%$, in the performance when facing the bottom flooding.

\begin{figure}[t]
	\label{fig:32x32}
	\centering
	\includegraphics[width=0.45\textwidth]{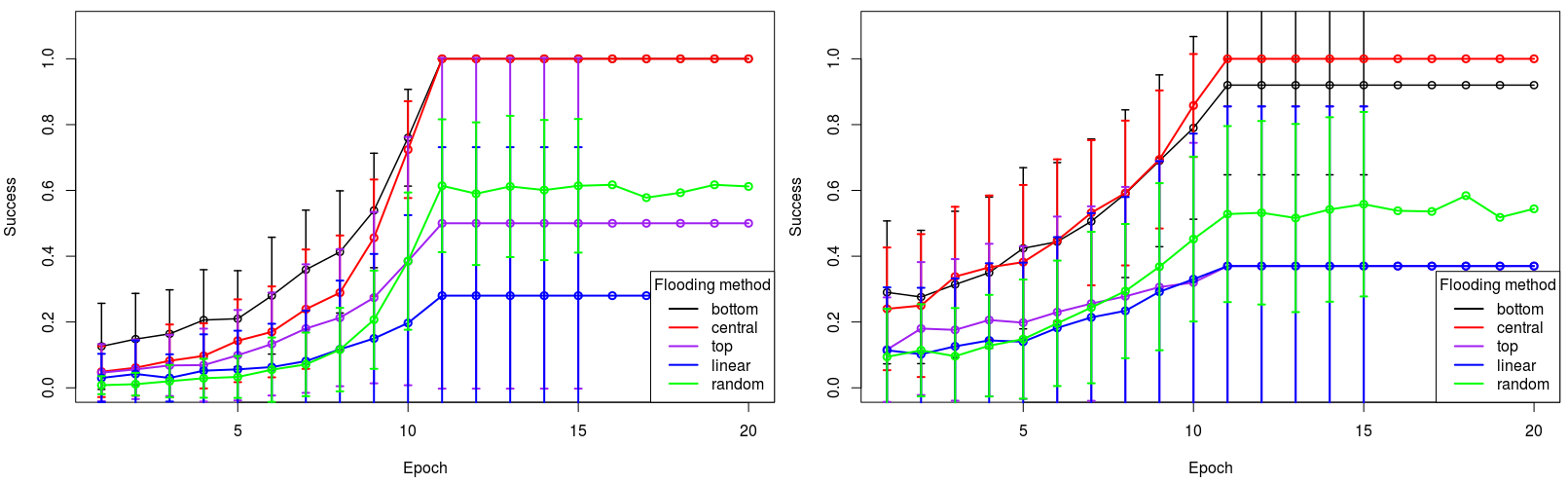}
	\caption{Flooding experiment Double Q-Learning convergence for the 32x32 maps with fewer objects, at the left, and the one with more objects, at the right.}
\end{figure}

The tests conducted with the $64\times64$ maps are portrayed in Figure \ref{fig:64x64}. It shows that the agent was able to obtain a $100\%$ success rate when facing the central and bottom floods, and performed the worst when facing the top and linear ones. On the dense version of this experiment, the was also observed a drop of $10\%$ in performance when facing the bottom flooding, and a $10\%$ increase when facing the linear one.

\begin{figure}[t]
	\label{fig:64x64}
	\centering
	\includegraphics[width=0.45\textwidth]{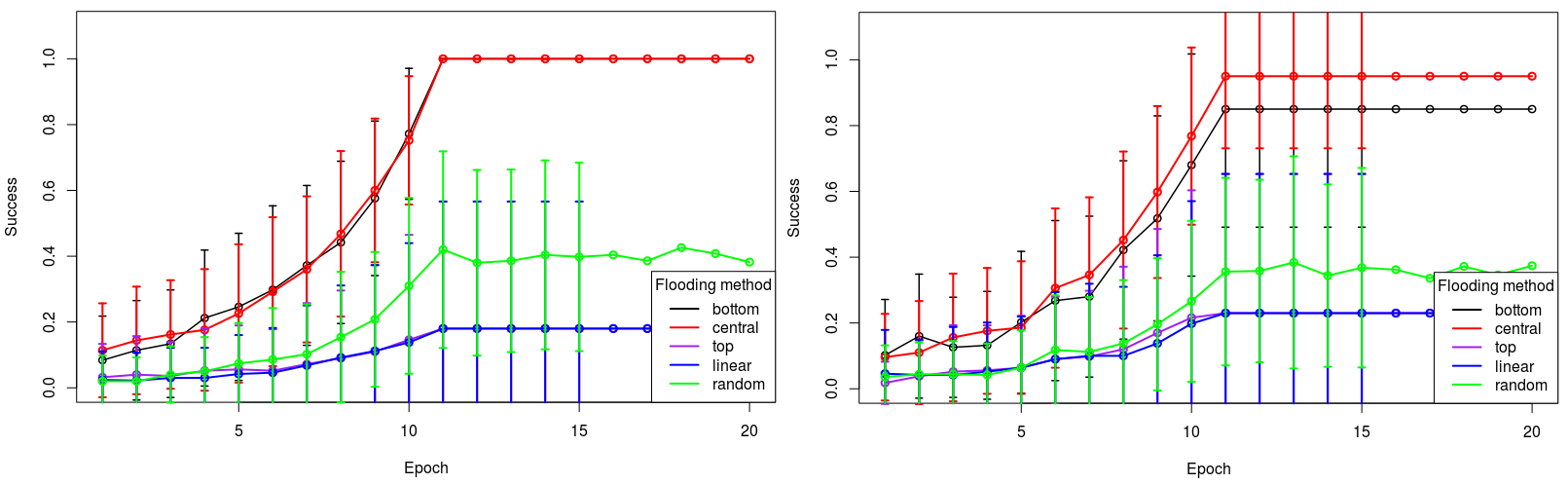}
	\caption{Flooding experiment Double Q-Learning convergence for the 64x64 maps with fewer objects, at the left, and the one with more objects, at the right.}
\end{figure}

The last experiment, conducted on the $128 \times 128$ maps, depicted in Figure \ref{fig:128x128}, shows that the agent was also able to obtain $100\%$ performance when facing both, the bottom, central, and top floods, and performed the worst when facing the linear and random ones. In contrast, for the dense versions of this map dimension, it was also observed a drop in almost $10\%$ of performance when performing under the simulation of the bottom flood. Moreover, it was also observed a performance decrease when facing the random ping flood. 

\begin{figure}[t]
	\label{fig:128x128}
	\centering
	\includegraphics[width=0.45\textwidth]{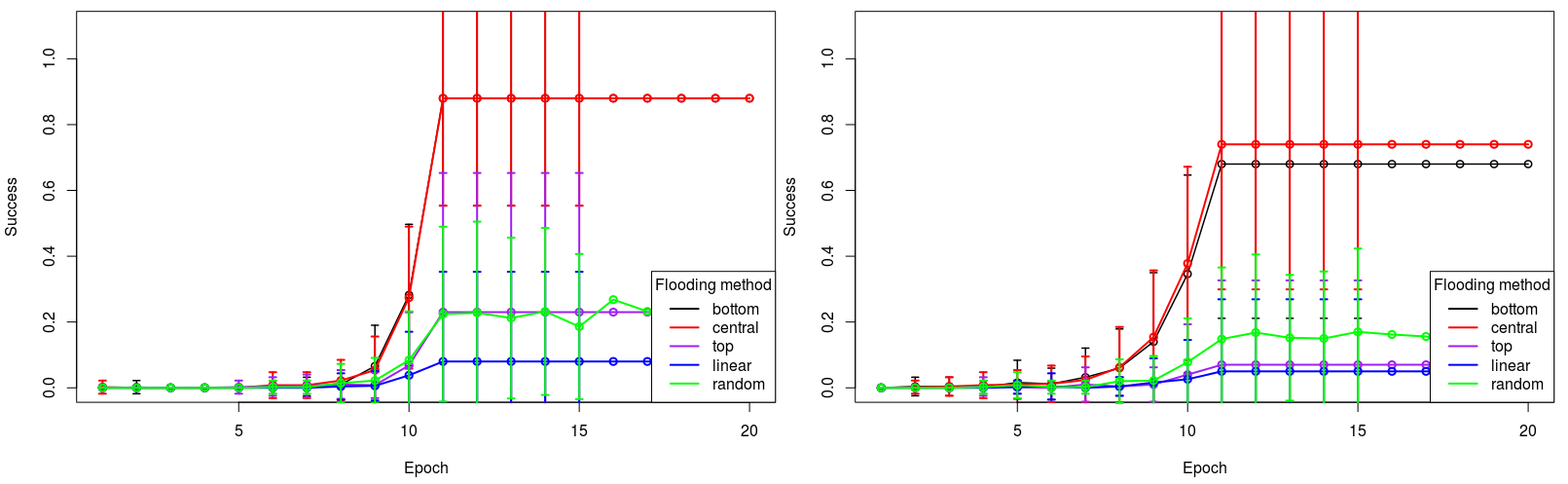}
	\caption{Flooding experiment Double Q-Learning convergence for the 128x128 maps with fewer objects, at the left, and the one with more objects, at the right.}
\end{figure}


\bibliographystyle{IEEEtran}
\bibliography{main}

\end{document}